\documentclass{article}
\usepackage[utf8]{inputenc}

\usepackage[preprint]{neurips_2020}
\usepackage{newtxtext}
\usepackage{amsmath}
\usepackage{graphicx}
\usepackage{amsfonts}
\usepackage{mathtools}
\usepackage{xcolor}
\usepackage{amsthm}
\usepackage{natbib}
\usepackage{hyperref}
\usepackage[ruled,vlined]{algorithm2e}
\usepackage{caption}
\usepackage{subcaption}
\usepackage{setspace}

\hypersetup{backref=true,       
    pagebackref=true,               
    hyperindex=true,                
    colorlinks=true,                
    breaklinks=true,                
    urlcolor= black,                
    linkcolor= blue,                
    bookmarks=true,                 
    bookmarksopen=false,
    filecolor=black,
    citecolor=blue,
    linkbordercolor=blue
}

\DeclareMathOperator*{\argmin}{argmin}

\DeclareMathOperator*{\Prb}{\mathbb{P}}

\theoremstyle{definition}

\title{Model Complexity of Program Phases}
\author{
  Arjun Karuvally\\
  \small College of Information and Computer Sciences \\
  \small University of Massachusetts Amherst \\
  \small Amherst, MA 01003 \\
  \texttt{akaruvally@cs.umass.edu} \\
  \And
  J. Eliot B. Moss \\
  \small College of Information and Computer Sciences \\
  \small University of Massachusetts Amherst \\
  \small Amherst, MA 01003 \\
  \texttt{moss@cs.umass.edu} \\
}

\begin{document}

\maketitle
\begin{spacing}{2}

\begin{abstract}

In resource limited computing systems, sequence prediction models must operate under tight constraints. 
Various models are available that cater to prediction under these conditions that in some way focus on reducing the cost of implementation. 
These resource constrained sequence prediction models in practice exhibit a fundamental tradeoff between the cost of implementation and the quality of its predictions.
This fundamental tradeoff seems to be largely unexplored for models for different tasks.
Here we formulate the necessary theory and an associated empirical procedure to explore this tradeoff space for a particular family of machine learning models such as deep neural networks. 
We anticipate that the knowledge of the behavior of this tradeoff may be beneficial in understanding the theoretical and practical limits of creation and deployment of models for resource constrained tasks.
\end{abstract}

\section{Introduction}

Sequence prediction models can have differing resource constraints imposed on them because of properties of the computing systems within which they are implemented.
%
Many of these models can be found in the design of computer systems where they are often used for improvements in performance of microarchitecture [\cite{Bhatia2019PerceptronBasedPF}, \cite{Calder1997EvidencebasedSB}, \cite{Dai2016Block2VecAD}], power/performance estimation [\cite{Carvalho2020UsingML}, \cite{Foots2020ClassificationOC}, \cite{Wu2015GPGPUPA}], thread/instruction scheduling [\cite{Li2009MachineLB}, \cite{Jain2019LearningAS}], and harware security [\cite{Chiappetta2016RealTD}, \cite{Khasawneh2017RHMDEH}, \cite{Khasawneh2020EnsembleHMDAH}].
%
%
%
There are two general scientific questions of interest in these models. 
The first one is: What is the nature of the tradeoff that exists between the cost to evaluate the model versus the quality of its predictions?
The second question is: What parts of a task require more complex models and how can we quantify and identify them?
We approach these questions from an algorithmic information theoretic perspective and explore a notion of information complexity that encompasses the properties of model implementation such as available data, model family, and the cost of evaluation of models.
%
%

We organize the paper as follows. In Section 1, we define sequence prediction problems with resource constraints. 
Section 2 defines our notion of information complexity based on Kolmogorov's algorithmic information theory, where we will also compare and contrast related notions that exist in the literature. 
In Section 3, we propose an empirical procedure to explore the nature of the proposed information complexity measure for resource constrained sequence prediction problems in general.
%
%
In Section 4, we show results from the empirical procedure applied to the problem of predicting future cache miss rates from the sequence of past cache miss rates. 
Miss rate prediction is useful for different problems in computer systems design like the optimization of energy consumption, cache partitioning, etc.

\section{Related Work}

Kolmogorov's notion of complexity has been used to define and quantify the complexity of objects from different domains such as genomics, virology, etc. 
\cite{Cilibrasi2005ClusteringBC} use Normalized Compression Distance (NCD), a metric based on Kolmogorov Complexity, to cluster identical objects. 
In NCD, a real world compressor is used to define the Kolmogorov Complexity of individual objects.
Recently, available data tends to be large and there is new interest emerging on questions about the relaxed or resource-bounded complexity of the data.
The general approach has been to find the complexity of the source generating the data and to quantify that complexity using the entropy of the source.
This was done by \cite{Rooij2012ApproximatingRG} who used standard lossy data compression techniques to find the trade-off between quality and representational cost.
In resource-bounded statistical models, we are really interested in the \textit{model} complexity of the data. This deviates from universal notions of complexity such as Shannon entropy.
\cite{10.1093/imaiai/iaaa033} used Kolmogorov Complexity of the model as a way to measure the difficulty of learning for a data set.
They validated this from the perspective of transfer learning by introducing a distance metric in the space of learning tasks.
This metric encompasses classical notions of Kolmogorov Complexity, and Shannon and Fisher Information.
From these works it can be observed that a general technique to make a computable version of the incomputable Kolmogorov Complexity is to use real compression methods.
In our work, we take ideas from the model compression literature to create an empirical procedure to explore the Kolmogorov complexity trade-off curves of cache miss rate behavior of computer programs.

Cache miss rate behavior analysis of programs is important for various applications.
\cite{Joshi2020MissRE} used a machine learning based method was to estimate cache miss rates for cache partitioning algorithm that optimizes cache utilization across multiple processors.
\cite{Srinivasan2013ProgramPD}, \cite{Khoshbakht2017ANA}, and \cite{Alcorta2021PhaseAwareCW} all used information about future program phases to improve the energy efficiency of a system. 
\cite{Chiu2018RuntimePP} used a run time method based on a decision tree model to create a mechanism to improve the energy delay product of programs running on a computer system with adjustable clock rates.
As machine learning based models are being deployed in computer systems, the question of the trade-off between cost of model evaluation and the quality of its predictions becomes important due to the resource constraints that exist in these systems.

\section{Resource Constrained Sequence Prediction Models} \label{section:rcspm_defn}

We now formalize sequence prediction problems and the associated models in which we are interested.
There are different ways to create models for sequence prediction problems. 
We focus on models representing Maximum Likelihood Estimates of a given sequence prediction task $\mathcal{T}$ under resource constraints.
In practice, the task $\mathcal{T}$ is defined using a finite data set $\mathcal{D} = \{ x_1, x_2, x_3 ... x_N \}$ that represents the sequence associated with the task.
A model learns the task using this data set and the data set is the only connection it can have to the task $\mathcal{T}$.
The maximum likelihood model for the task $\mathcal{T}$ is represented by a computable probability distribution $f$ with parameters $\theta$. 
That is, $f$ is a predictor that offers a \textit{distribution} of values of the next elements in the sequence as opposed to predicting a single most-likely value.
The negative log likelihood ($\mathcal{L}_{\mathcal{D}}$) of the model $f$ with respect to the data set $\mathcal{D} = \{ x_1, x_2 ... x_N  \}$ serves as a measure of fit of the model to the task:
\begin{equation}
    \mathcal{L}_{\mathcal{D}}(f_\theta) = \sum_{i=2}^{N} - \log(f_{\theta}(x_i|x_1, x_2 ... x_{i-1}))
\end{equation}
The cost of evaluation of the model with parameters $\theta$ is given by a function $J(\theta)  \geq 0$. 
%
%
%
The resource constraints that are imposed on the model are given by $\alpha$ which constraints $J$ such that $J(\theta) \leq \alpha$. 
Using the above definitions, we formalize the resource-constrained sequence prediction problem as that of finding parameter vector $\theta^*$ that minimizes loss, subject to the resource constraint:
\begin{equation}
    \theta^* = \argmin_{\theta} \mathcal{L}_{\mathcal{D}}(f_\theta), \text{such that } 0 \leq J(\theta) \leq \alpha
\end{equation}

We assume in this work that the model $f$ is a computable probability distribution in the space of all probability distributions. 
This assumption seems reasonable since most models built for these problems run in a Turing Machine (any computer system) that can represent only computable functions $f$.

\begin{figure}	
	\begin{subfigure}[t]{1in}
        \hspace*{2cm}
		\includegraphics[width=4in]{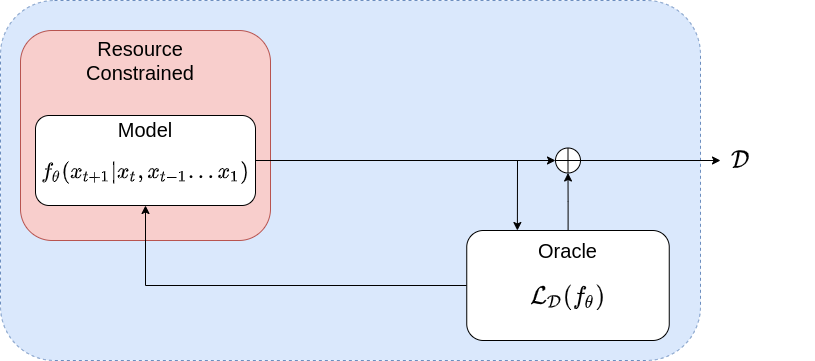}
	\end{subfigure}
	\caption{The general architecture of resource constrained sequence prediction models. The data set $\mathcal{D}$ is generated using two parts. A model $f_\theta$ that has resource constraints imposed on it and an associated oracle that helps the model encode $\mathcal{D}$ using $\mathcal{L}_{\mathcal{D}}(f_{\theta})$, the losses, from which one can derive the errors and thus correct the predictions.}\label{fig:oracle_architecture}
\end{figure}

%
Using these definitions, the sequence prediction model with which we are concerned with in this paper has the general architecture shown in Figure \ref{fig:oracle_architecture}.
The sequence prediction model forms an auto-regressive feedback loop with an oracle that corrects the predictions of the model and provides the model with information about the quality of its predictions at any point in time $t$. 
The resource constraints are imposed on the model part of the system. 
Although the oracle is not explicitly resource constrained, a tradeoff exists between the size of the oracle and the complexity of the model, which is given by the structure function of the task $\mathcal{T}$. We will elaborate on this formally in Section \ref{sec:structure_function}.

\section{Algorithmic Information Content} \label{sec:structure_function}
\begin{figure}	
	\begin{subfigure}[t]{1in}
        \hspace*{3cm}
		\includegraphics[width=2.5in]{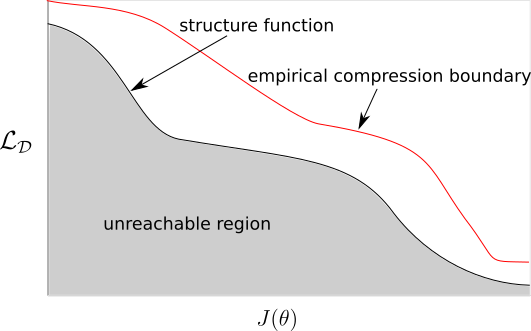}
	\end{subfigure}
	\caption{The tradeoff between the cost of implementation for a model to the loss associated with its predictions. The structure function marks the boundary of the set of feasible models. The empirical compression boundary that we explore in this paper is an upper bound to this structure function.}\label{fig:hypothetical_tradeoff}
\end{figure}
As the cost to evaluate a model $f_{\theta}$ is related to the description of the model with respect to a finite data set $\mathcal{D}$, we leverage ideas from algorithmic information theory to define a notion of information content of interest in our context.
A principled notion of information content defined for the type of problems formalized in Section \ref{section:rcspm_defn} should encompass choices such as the model family, finite data set, and the cost function of interest.
%
Algorithmic information theory defines the information content of a string $x$ as its Kolmogorov Complexity. 
The Kolmogorov Complexity of a string $x$ is defined as the shortest possible description in a program written in a universal language $U$, where the program outputs $x$.
%
%
Formally, consider $p$ to be any program that can be run on a universal machine $U$ where $p$ outputs $x$ and $l(p)$ is the length of the program $p$. The Kolmogorov Complexity for a string $x$ is defined as
\begin{equation}
    K_{U}(x) = \min_{p} \bigl\{ l(p): U(p) = x \bigr\}
\end{equation}
%
Using Kolmogorov Complexity, the amount of information content in the data set $\mathcal{D}$ can thus be defined as
\begin{equation}
    K_{U}(\mathcal{D}) = \min_{p} \bigl\{ l(p): U(p) = \mathcal{D} \bigr\}
\end{equation}
To define the information content of interest for the  resource constrained sequence prediction problems of Section \ref{section:rcspm_defn}, we first consider the following notion of information content associated with the two part description of the data set $\mathcal{D}$ using the model $f_{\theta}$. 
This will be the combined description of the model along with the oracle in the general architecture.
\begin{equation}
    C_K(\mathcal{D}, f_{\theta}) = \min_{\theta} \bigl\{ \mathcal{L}_{\mathcal{D}}(f_\theta) + K_U(f_\theta) \bigr\}
\end{equation}
One part of the description $C_K$ is $K_U(f_\theta)$ which is the complexity of description of the model $f_{\theta}$ in a universal language $U$. 
The other part, $\mathcal{L}_{\mathcal{D}}(\theta)$, is the complexity of encoding the data set $\mathcal{D}$ using the model $f_{\theta}$. 
Since after learning the data set $\mathcal{D}$, only $\theta$ is left in the description of the model, the set of all $\theta$ forms statistics of the model with respect to the data set $\mathcal{D}$.
The setting of $\theta$ that satisfies $C_K(\theta)$ forms both the sufficient and minimal statistics of the model with respect to data set $\mathcal{D}$ \citep{Vereshchagin2004KolmogorovsSF}.
%
%
%
This definition of $C_K$ does not incorporate the notion of resource constraints and the tradeoff between $K_U(f_{\theta})$ and $\mathcal{L}_{\mathcal{D}}(f_{\theta})$.
To allow us to reason about the tradeoff, we relax the notion of the minimal sufficient statistic $\theta$ and consider statistics that are only $\beta$ sufficient.
$\beta$ sufficient statistics are the statistics $\theta$ that satisfy the following relaxed notion of information content:
\begin{equation}
    C_{\beta}(\mathcal{D}) = \min_{\theta} \mathcal{L}_{\mathcal{D}}(f_\theta) + \beta K_U(f_\theta)
    \label{eqn:beta_complexity}
\end{equation}
It was shown by \cite{Vereshchagin2004KolmogorovsSF} that for $\beta = 1$, the statistic $\theta$ that satisfies $C_{\beta}$ matches Kolmogorov's notion of minimal sufficient statistics and relates to the phase change in $\theta$ that occurs when a model enters the overfitting regime for a given data set $\mathcal{D}$.
This draws a connection between resource constrained Kolmogorov Complexity and the theory of learning. 
Since our work is mainly concerned with the tradeoff suggested by $C_{\beta}$, we do not delve into the implications of it in learning theory.
The tradeoff between $J(\theta)$ and $\mathcal{L}_{\mathcal{D}}(f_{\theta})$ inferred by $C_{\beta}$ is given by its associated structure function which is defined as

\begin{equation}
    S_{\mathcal{D}}(\alpha) = \min_{\theta} \bigl\{ \mathcal{L}_{\mathcal{D}}(f_\theta) : K_{U}(f_\theta) \leq \alpha \bigr\}
    \label{eqn:structure_function_beta}
\end{equation}

Thus the value of $\beta$ in $C_{\beta}$ serves as a Lagrange multiplier of the structure function in Equation \ref{eqn:structure_function_beta} and is related to different resource constraints imposed on the model.
Now that we have a notion of resource constraint in $C_{\beta}$, we focus on incorporating the cost function into $C_{\beta}$. 
Since evaluation cost can be well approximated by the description of the model $f_\theta$, we incorporate a notion of cost into the universal language $U$. 
Consider $K_J(f_\theta)$ to be the cost to encode $f_{\theta}$ using a cost model $J$.
%
We fix $J$ and the relation between $K_{U}$ such that

\begin{equation}
    K_U(f_\theta) + c \leq K_J(f_\theta)
\end{equation}

Here $c$ is the length of a fixed size "shell" program independent of the input that encodes the cost function $J$. 
The cost function serves as an upper bound to the algorithmic information content $K_U(f_{\theta})$.
The structure function that incorporates our notion of cost can be defined as:

\begin{equation}
    S_{J, \mathcal{D}}(k) = \min_{\theta} \bigl\{ \mathcal{L}_{\mathcal{D}}(f_{\theta}) : K_{J}(f_\theta) \leq k \bigr\}
\end{equation}

%
The structure function $S_{J, \mathcal{D}}(k)$ reformulates the search for a model for a task as a problem of encoding the structure and variability present in a data set $\mathcal{D}$.
Ideally, the structure that should be learned for solving a task should be encoded using the description of the model $K_{J}$ and the nuisance variability is encoded using $\mathcal{L}_{\mathcal{D}}$.
The tradeoff between the descriptions $K_J$ and $\mathcal{L}_{\mathcal{D}}$ defines the nature of the fundamental tradeoff between the cost of evaluation of a model and the quality of its predictions for the resource constrained sequence prediction problem defined in Section \ref{section:rcspm_defn}. 

\section{Empirical Compression Boundary} \label{section:empirical_compression_boundary}


The structure function of a learning task provides a theoretically grounded framework to answer questions about the tradeoff between cost and quality of a model.
Unfortunately, quantities based on Kolmogorov Complexity are not computable even in theory and the notion works only in asymptotic regimes. 
Hence the exact nature of the function is not discernable for all possible model families unless the entire parameter space for each model in the model families can be explored.
Exploration of the entire parameter space is intractable for deep neural networks where the number of parameters is very high.
%
We leverage ideas from the model compression literature to develop an empirical procedure to explore an upper bound to this function which we call the \textit{compression boundary}.
To simplify the development of an empirical framework, we fix the cost function of interest to be the number of non-zero parameters of the model.
\begin{equation}
    J(\theta) = \sum_{i} |\theta_i| \,\,\, \text{such that} \,\,\,\, |x| = \begin{cases}
        1 & x \neq 0 \\
        0 & x = 0
    \end{cases}
\end{equation}
This cost function does not consider the full description length of a model. For example, additional cost may be required to encode each of the parameters $\theta$.
For a model that can have at most $n$ parameters, the exact description can have a complexity given by $K_J(f_\theta) \leq a J(\theta) + b \log(n) + c$.
This description consists of a three components. The first component $a J(\theta)$ is the cost to encode all the parameters. Here, $a$ is the number of bits to encode each $\theta$ on average.
The second component, $b \log(n)$, is the cost to encode the \emph{position} of each $\theta$, needed for describing the compressed representation.
$b$ is a multiplier that allows for description of other parameter position related information such as activation functions, layer types, etc.
The last description component is a constant $c$ that is independent of the number of parameters.

For settings of $\theta$ where $J(\theta) \gg 1$ (asymptotic regimes), the effect of these additional components can be ignored when comparing different boundaries.
%
This condition seems to be satisfied for the deep learning models with very many parameters with which we are concerned here.
%
We can define an optimization problem to explore the compression boundary using convex optimization procedures. Formally, the boundary is defined by all $\theta^*$ that satisfy
\begin{equation}
    \theta^* = \arg \min_{\theta} \mathcal{L}_{\mathcal{D}}(\theta) + \beta \sum_{i} |\theta_i|
    \label{eqn:compression_boundary}
\end{equation}
Since we use convex optimization procedures with gradient descent to learn the parameters $\theta$ for a model, it is difficult to back-propagate through the discrete function $|x|$.
This issue can be alleviated by a variety of methods. 
%
We have found in our experiments that the exact method used to optimize this function does not have much influence on finding the compression boundary as all methods lead to the same boundary.
We choose to solve the optimization problem in Equation \ref{eqn:compression_boundary} by introducing another set of parameters to the model called gates ($g$) defined as $g = \text{sigmoid}(z)$. $\theta$ is redefined using $g$ as $\theta' = g \cdot \theta$. The optimization procedure is then reformulated as
\begin{equation}
    \theta^* = \arg \min_{\theta'} \mathcal{L}_{\mathcal{D}}(\theta') + \beta \sum_{i} g
\end{equation}
$g$ acts as a smooth approximation to the discrete function $|\theta|$ because $0 < g < 1$. 
Since $g$ cannot be exactly zero due to the sigmoid function, we introduce a magnitude based threshold, $g_{\min}$, for each of the gates, and consider $\theta_i$ to be $0$ if $g_i \leq g_{\min}$.
We propose to leverage the Legrange multiplier $\beta$ and the magnitude based threshold $\alpha$ to find the compression boundary.
We present the algorithm to explore the tradeoff space in Algorithm \ref{algo:compression_boundary}. 

\begin{algorithm}[t]
\SetAlgoLined
Fix range of beta - [$\beta_0$, $\beta_1$] \;
Fix magnitude based threshold - [$g'_{\min}, g'_{\max}$] \;
Fix ordered set of $\beta$ to explore - $\{ \beta'_0, \beta'_1 .., \beta'_i, ... \}$ where $\beta_0 \leq \beta'_i < \beta'_{i+1} \leq \beta_1$ \;
Fix ordered set of $g_{\min}$ to explore - $\{ g'_0, g'_1 .., g'_i, ... \}$ where $g'_{\min} \leq g'_i < g'_{i+1} \leq g'_{\max}$ \;
\ForEach{$\beta \in \{ \beta'_0, \beta'_1 .., \beta'_i, ... \}$ \;}{
Solve $\theta = \argmin_{\theta} \mathcal{L}_{\mathcal{D}}(\theta) + \beta \sum_{i} |\theta_i|$ \;
\ForEach{$g_{\min} \in \{ g'_0, g'_1 .., g'_i, ... \}$}{
    $\forall i : g_i < g_{\min} :$ set $ g_i \leftarrow 0$ \;
    Recompute $\forall i: \theta_i := g_i \cdot \theta_i$ \;
    Extract Model $\mathcal{H}(\beta, \alpha)$ \;
}
}
 \caption{Compression Boundary Estimation}
 \label{algo:compression_boundary}
\end{algorithm}

\section{The Cache Miss Rate Prediction Problem}

\begin{figure}[!tbp]
  \begin{minipage}[b]{0.98\textwidth}
    \begin{subfigure}{0.32\textwidth}
        \includegraphics[width=0.95\linewidth]{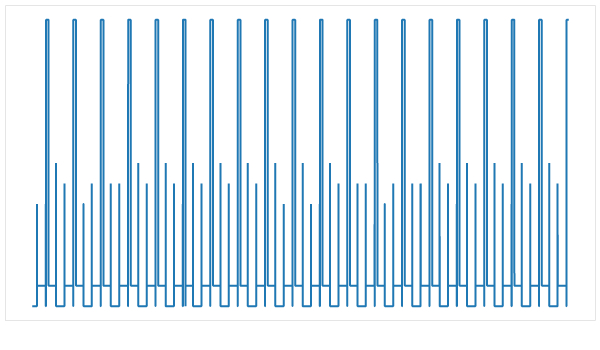}
        \caption{swim-ref-swim}
        \label{fig:cmr:swim-ref-swim}
    \end{subfigure}
    \begin{subfigure}{0.32\textwidth}
        \includegraphics[width=0.95\linewidth]{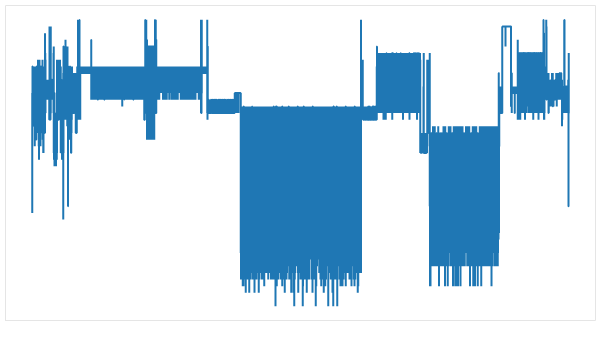}
        \caption{gcc-ref-cc15}
        \label{fig:cmr:gcc-ref-cc15}
    \end{subfigure}
    \begin{subfigure}{0.32\textwidth}
        \includegraphics[width=0.95\linewidth]{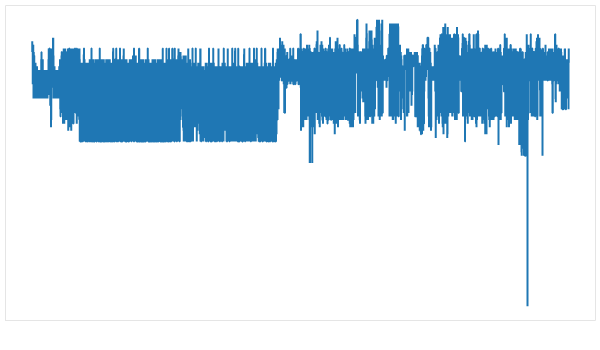}
        \caption{pmd-small-JikesRVM}
        \label{fig:cmr:pmd-small-JikesRVM}
    \end{subfigure}
  \end{minipage}
  \caption{$\log_{10}$ of cache miss rates over time for programs with three different behaviors depending on the length of phases they exhibit.}
  \label{fig:cachemissrates}
\end{figure}

In the cache miss rate prediction problem, we wish to predict the miss rate of a cache over time encountered by a program running on a CPU.
The sequence prediction model predicts the next miss rate given a sequence of previous miss rates.
Most computing systems on which these models are implemented impose space and time constraints on the model if it is to be used in real time.
Thus in these models, the procedure that we developed in Section \ref{section:rcspm_defn} can be used to explore the associated cost-quality tradeoff.
It is difficult to build a very general model for cache miss rate prediction as programs running in a computer system can vary significantly in their behavior, which is evident from Figure \ref{fig:cachemissrates}.
Here we focus instead on the cache miss rate behavior of individual programs that have different phases [\cite{Chiu2018RuntimePP}] during their execution, and we compare the associated tradeoff curves. 
We chose three different programs to explore, selected based on the number and duration of their phases of execution. 
For swim-ref-swim, shown in Figure \ref{fig:cmr:swim-ref-swim}, the behavior of the cache miss rate is fairly regular throughout the trace, which points to a single long phase.
For gcc-ref-cc15, shown in Figure \ref{fig:cmr:gcc-ref-cc15}, there are more distinct phases but each phase is shorter.
pmd-small-JikesRVM contrasts with the other two traces because the cache miss rate behavior does not show as much visual evidence of phase behavior, which could mean that the phases are transient (short).
This trace thus depicts the case where a model may have trouble finding a regular pattern across the trace.
%
From the perspective of building a resource constrained model, we expect the cost to encode the dynamic behavior of the above programs to follow the trend:  swim-ref-swim < gcc-ref-cc15 < pmd-small-JikesRVM.
%
%

\section{Experimental Setup}

\begin{figure}[!tbp]
  \begin{minipage}[b]{0.98\textwidth}
    \begin{subfigure}{0.5\textwidth}
        \includegraphics[width=0.95\linewidth]{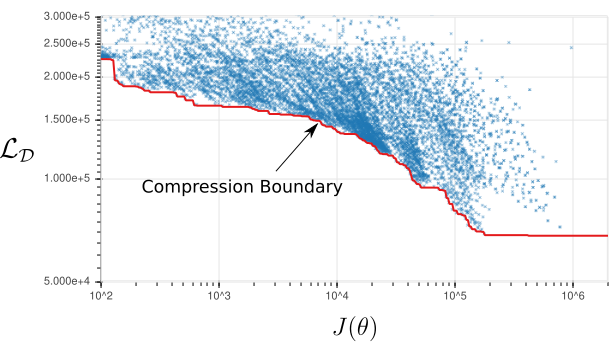}
        \caption{}
        \label{fig:compression_boundary_gcc}
    \end{subfigure}
    \begin{subfigure}{0.5\textwidth}
        \includegraphics[width=0.95\linewidth]{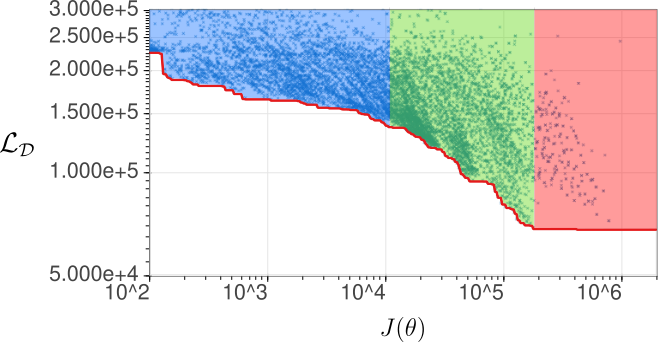}
        \caption{}
        \label{fig:theta_phases_gcc}
    \end{subfigure}
  \end{minipage}
  \caption{Compression boundary and parameter phases for gcc-ref-cc15 with LSTM models. Each blue point corresponds to a model $\mathcal{H}(\beta, \alpha)$ found by the Empirical Compression Boundary algorithm.}
\end{figure}

\begin{figure}[!tbp]
  \begin{minipage}[b]{0.98\textwidth}
  \centering
    \begin{subfigure}{0.64\textwidth}
        \includegraphics[width=0.95\linewidth]{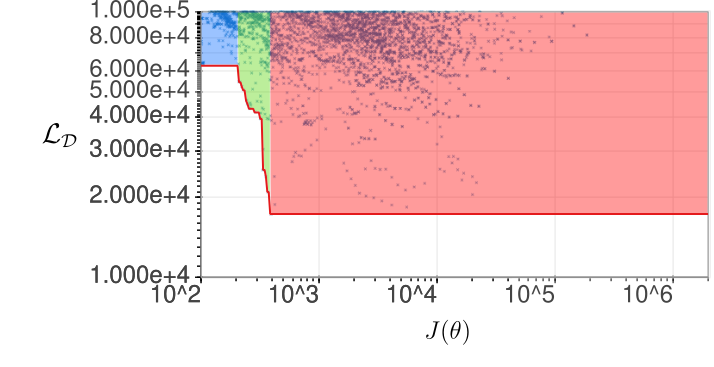}
        \caption{swim-ref-swim}
        \label{fig:theta_phases_swim}
    \end{subfigure} \\
    \vspace{20px}
    \begin{subfigure}{0.64\textwidth}
        \includegraphics[width=0.95\linewidth]{figs/gcc_phases.png}
        \caption{gcc-ref-cc15}
        \label{fig:theta_phases_gcc}
    \end{subfigure} \\
    \vspace{20px}
    \begin{subfigure}{0.64\textwidth}
        \includegraphics[width=0.95\linewidth]{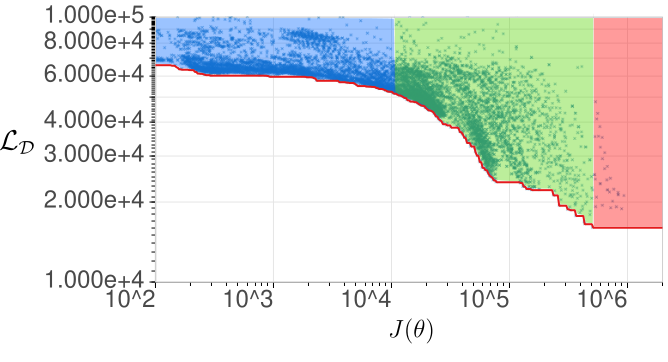}
        \caption{pmd-small-JikesRVM}
        \label{fig:theta_phases_pmd}
    \end{subfigure}
  \end{minipage}
  \caption{Compression boundaries and parameter phases of $\beta$ sufficient parameters of LSTM for the three traces. Models with number of parameters from $10$ to $2 \times 10^6$ are explored.
  }
  \label{fig:parameter_phases:comparison}
\end{figure}

\textbf{Data Collection:}
gcc-ref-cc15 and swim-ref-swim are part of the SPEC CPU 2000 benchmark suire (“ref” size runs) where we collected traces of every memory access made by the programs using valgrind, specifically its Lackey tool. 
We used the same tool on pmd-small-JikesRVM which is part of the DaCapo Java benchmark suite. 
We mapped virtual addresses of data accesses to their 64-byte virtual
cache line, and applied the least recently used (LRU) stack algorithm to obtain miss rates for various cache sizes of perfect LRU caches. 
These are calculated over windows of 100,000 instructions.
%

\textbf{Data Preprocessing:}
We transform the cache miss rates using $\log_{10}$ to show better the
interesting miss rates close to $0$.  
To avoid $0$ itself, we clip the lower bound of the value of miss rates to have at least some small value $0 < \epsilon < 1/100000$. 
For learning, the sequences are divided into contiguous chunks, with some chunks used for training and others for testing.
Training and test chunks are drawn from all parts of the traces so
that all different behaviors within a program are captured in both training and testing.

\textbf{Probabilistic Modeling:}

We take the time evolution of the preprocessed cache miss rates of a program as the data set $\mathcal{D} = \{ x_1, x_2 ... x_N \}$ to learn. 
Since each $x_i$ is a real number and not a discrete quantity, it is not ideal for modeling as a sequence.
We bin the $x_i$ into 100 equally spaced bins so that we can discretize the space making it suitable to perform sequence modeling [\cite{Salman2012RegressionAC}].
Thus, each $x_i$, which was a real number, is converted to $x'_i$ which is a discrete integer value in $[0, 99]$.
The data set for learning is changed to be $\mathcal{D} = \{ x'_1, x'_2 ... x'_N \}$. 
%
The objective of the model is now to estimate the distribution
\begin{equation}
    \Prb(x'_{t} | x'_1, x'_2 ... x'_{t-1} )
\end{equation}
Now we can apply the techniques we devised in Section \ref{section:empirical_compression_boundary} to explore the cost-quality tradeoff curves of the three benchmark traces.
We used LSTMs \citep{Hochreiter1997LongSM} as the model family under consideration.
LSTMs are a variant of RNNs \citep{Rumelhart1986LearningIR} and have been
successful in sequence modeling due to their ability to capture short
and long term dependencies in sequential data [\cite{Sutskever2014SequenceTS,
Wu2016GooglesNM}].  
%
%
%
\emph{Unrolling} LSTMs beyond a
certain time step in the history of a sequence leads to heavy computation and
\emph{vanishing gradient} issues
[\cite{vanishinggradient,vanishinggradientbengio}], so we modified the model to
account for this by unrolling only up to a finite number of time steps, $h$, behind
the current history of the sequence.  Our model consists of an LSTM layer
followed by 4 feed forward layers [\cite{lstmmodel}].  The hidden state
of the LSTM is forwarded on to the next prediction of the model.  The hidden
state, in theory, contains information about the values prior to the current
time step.

\begin{figure}	
	\begin{subfigure}[t]{1in}
        \hspace*{3cm}
		\includegraphics[width=2.5in]{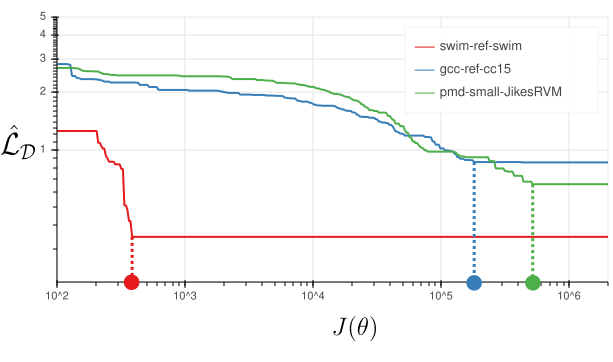}
	\end{subfigure}
	\caption{Tradeoff between the average negative log likelihood ($\hat{\mathcal{L}}_{\mathcal{D}}$) and the cost ($J(\theta)$) for all three traces. It is observed that swim-ref-swim has the lowest cost at the point at which the parameter phase change happens, followed by gcc-ref-cc15 and pmd-small-JikesRVM in that order.}
	\label{fig:parameter_phase:average_loss}
\end{figure}

We run the Empirical Compression Boundary algorithm on the data sets $\mathcal{D}$ generated by the three programs. 
The models to be learned are initialized with different seeds and layer widths to get better estimates of the boundary.
We plot the loss $\mathcal{L}_{\mathcal{D}}$ against the cost $J(\theta)$ for all $\mathcal{H}$ obtained by running the compression boundary algorithm.
%
%


\section{Results} \label{section:results}

\begin{figure}[!tbp]
  \begin{minipage}[b]{0.98\textwidth}
    \begin{subfigure}{0.5\textwidth}
        \includegraphics[width=0.95\linewidth]{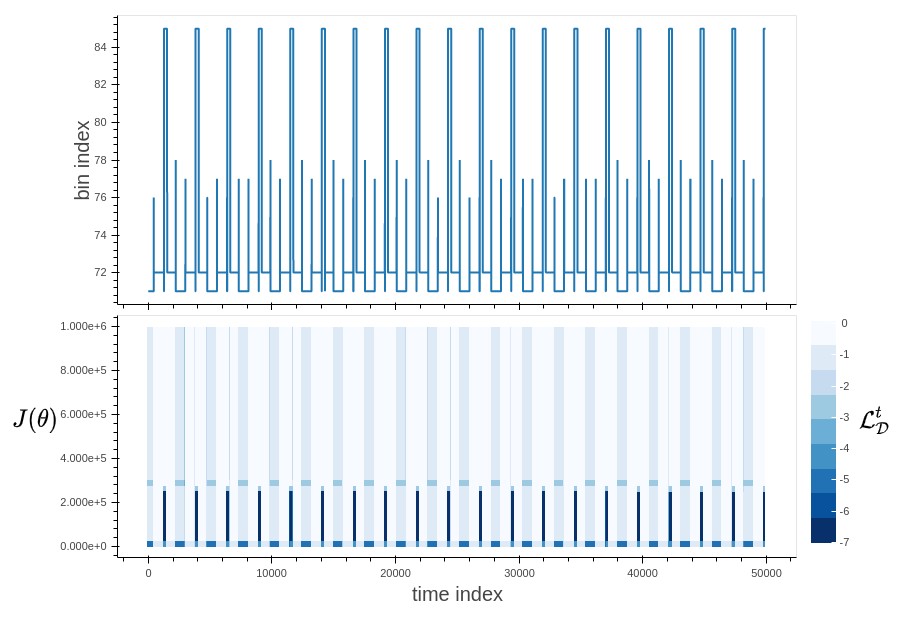}
        \caption{}
        \label{fig:ll_heatmap:swim}
    \end{subfigure}
    \begin{subfigure}{0.5\textwidth}
        \includegraphics[width=0.95\linewidth]{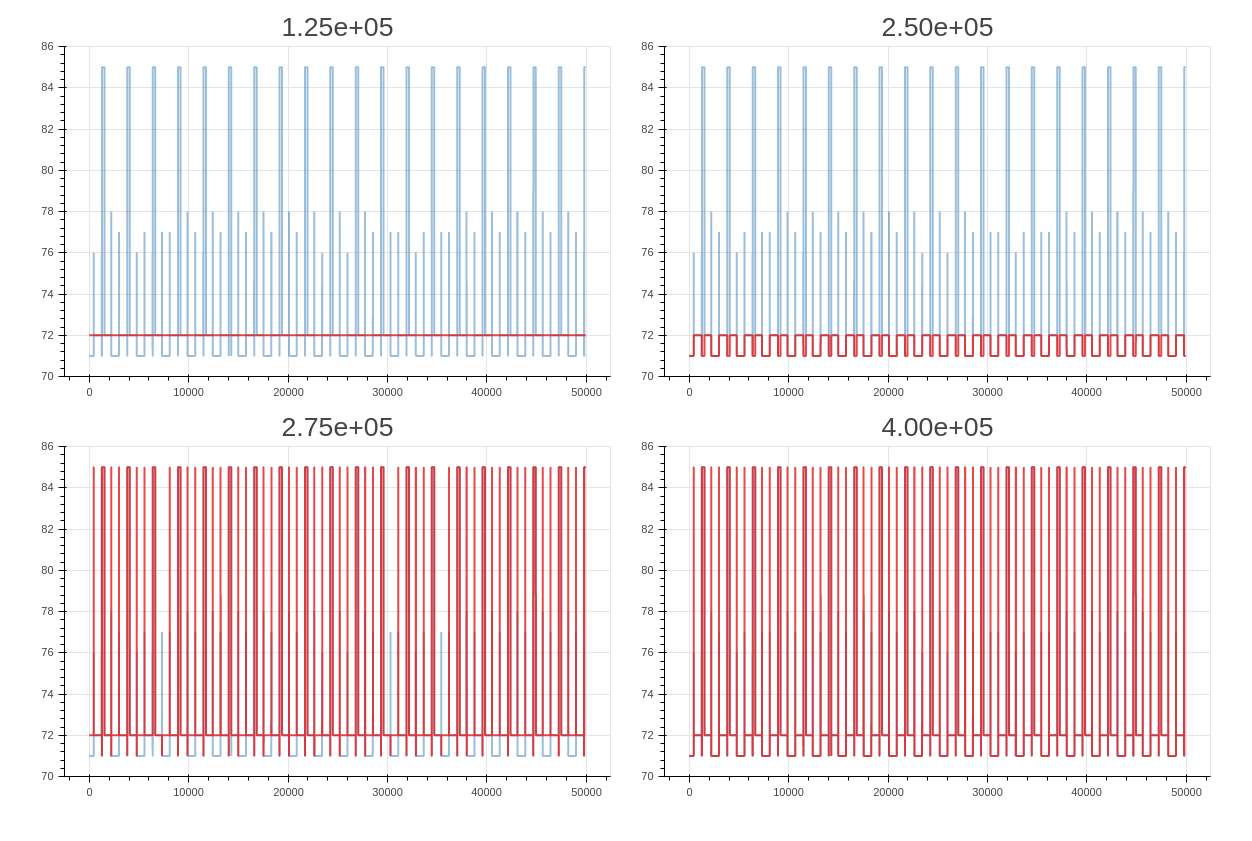}
        \caption{}
        \label{fig:cost_predictions:swim}
    \end{subfigure}
  \end{minipage}
  \caption{Local likelihoods and predictions for models of swim-ref-swim. Figure \ref{fig:ll_heatmap:swim} shows the local likelihoods across time for models with different cost constraints $J(\theta)$. Figure \ref{fig:cost_predictions:swim} shows the predictions of models for different costs. The blue line represents the actual value of the bin at every point in the trace and red lines correspond to the predictions. The improvement of predictions across the trace can be observed as the cost constraints are relaxed. 
  }
  \label{fig:local_likelihood:swim}
\end{figure}

\begin{figure}[!tbp]
  \begin{minipage}[b]{0.98\textwidth}
    \begin{subfigure}{0.5\textwidth}
        \includegraphics[width=0.95\linewidth]{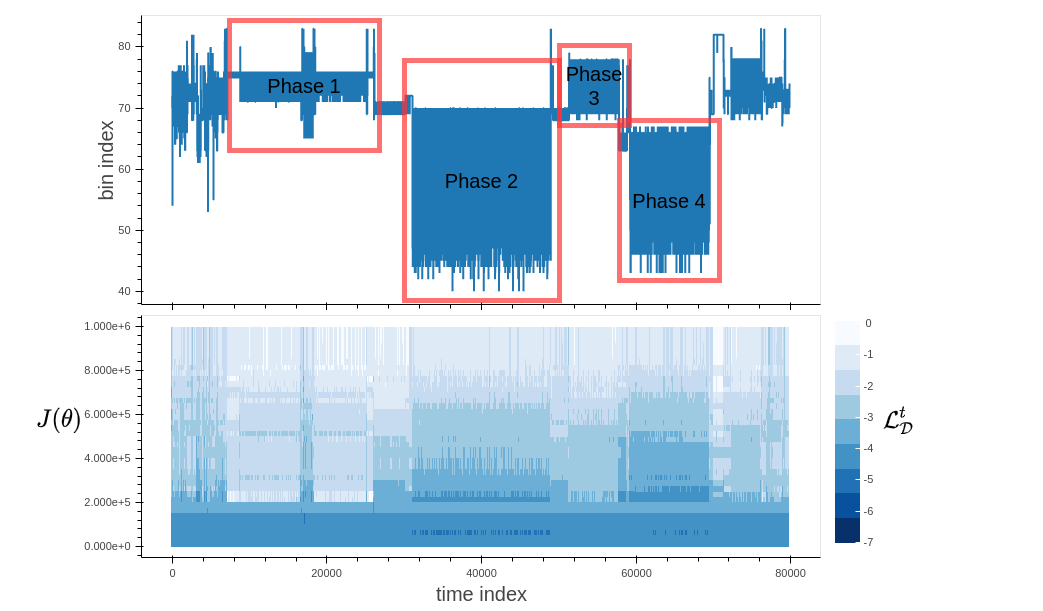}
        \caption{}
        \label{fig:ll_heatmap:gcc}
    \end{subfigure}
    \begin{subfigure}{0.5\textwidth}
        \includegraphics[width=0.95\linewidth]{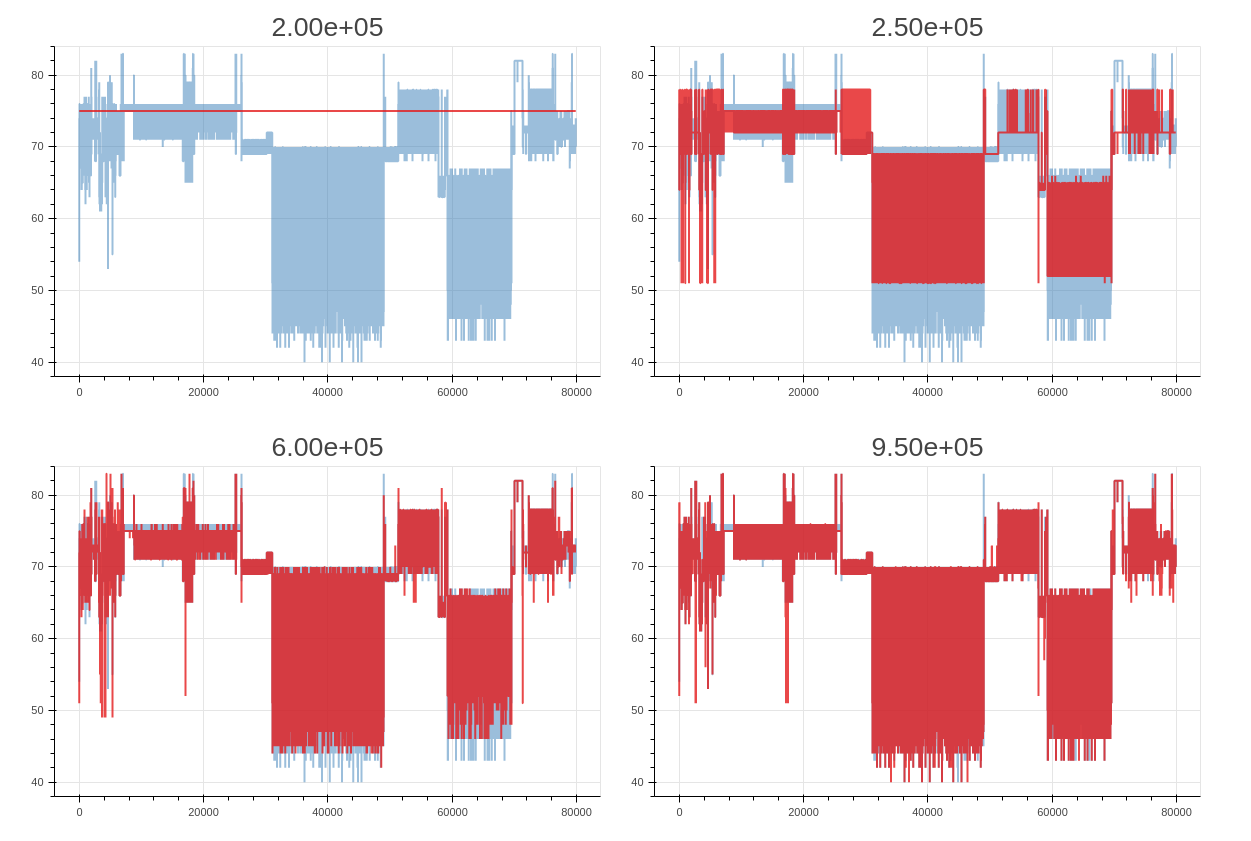}
        \caption{}
        \label{fig:cost_predictions:gcc}
    \end{subfigure}
  \end{minipage}
  \caption{Local likelihoods and predictions for models of gcc-ref-cc15.}
  \label{fig:local_likelihood:gcc}
\end{figure}

\begin{figure}[!tbp]
  \begin{minipage}[b]{0.98\textwidth}
    \begin{subfigure}{0.5\textwidth}
        \includegraphics[width=0.95\linewidth]{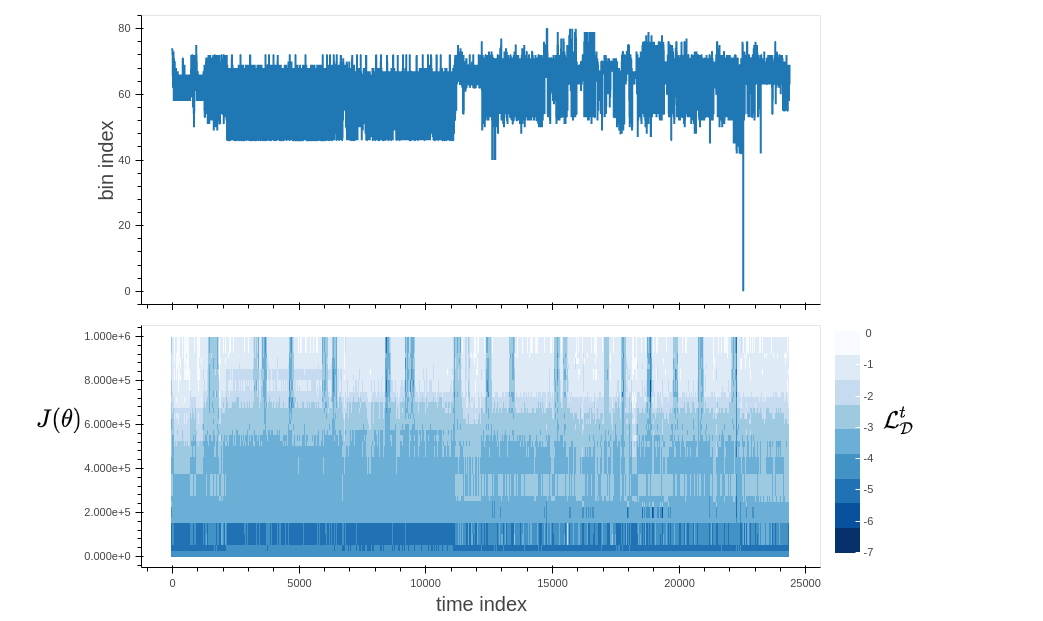}
        \caption{}
        \label{fig:ll_heatmap:pmd}
    \end{subfigure}
    \begin{subfigure}{0.5\textwidth}
        \includegraphics[width=0.95\linewidth]{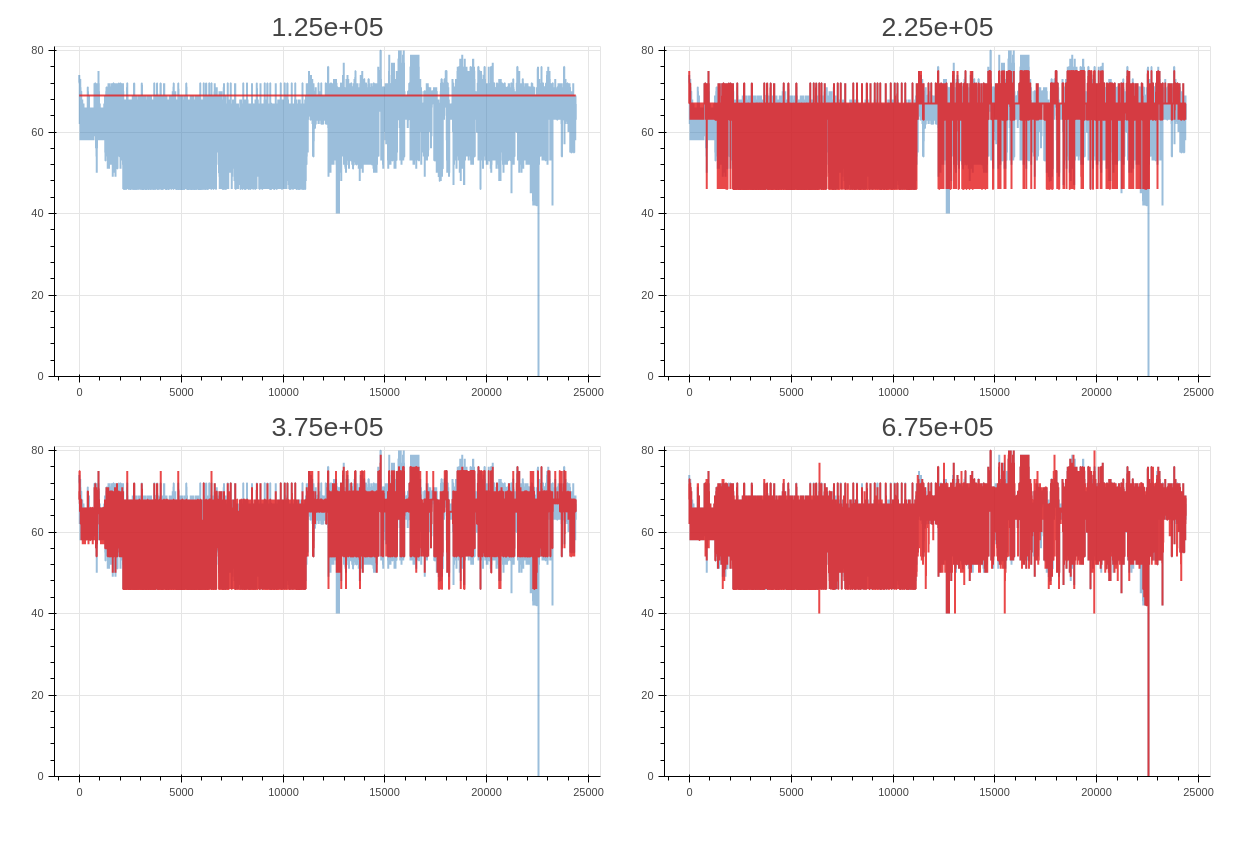}
        \caption{}
        \label{fig:cost_predictions:pmd}
    \end{subfigure}
  \end{minipage}
  \caption{Local likelihoods and predictions for models of pmd-small-JikesRVM.
  }
  \label{fig:local_likelihoods:pmd}
\end{figure}

%
We tackle the first question: What is the nature of the tradeoff that exists between the cost to evaluate the model to the quality of its predictions? We try to answer this question by looking at the general behavior of the compression boundary for each of the traces that we selected. 
The range of the cost $J(\theta)$ is fixed between $10$ and $2 \times 10^6$ since the tradeoff region of the three traces occurs in this range.
Although we will focus the following discussion on the trace generated by gcc-ref-cc15, the general behavior we analyze is valid for all the other traces.
Figure \ref{fig:compression_boundary_gcc} shows the tradeoff between the loss $\mathcal{L}_{\mathcal{D}}$ and cost $J(\theta)$ obtained for gcc-ref-cc15.
%
The models along the compression boundary can be divided as belonging to three different phases based on the slope at each point. 
To clarify the difference between phases discussed here and phases in behavior of cache miss rate traces that was mentioned before, we will use \textit{parameter phase} to mean the phase of the parameter of the model and we will use \textit{program phase} to mean the phase-like behavior of the program. 
Three \textit{parameter phases} can be observed in the log-log scale:
\begin{itemize}
    \item \textbf{Phase 1} (blue region in Figure \ref{fig:theta_phases_gcc}) - The region of model parameters where we have to invest significantly in the cost of the model to obtain small factor improvements in the quality of its predictions.
    \item \textbf{Phase 2} (green region) - The region of model parameters where the tradeoff between $J(\theta)$ and $\mathcal{L}_{\mathcal{D}}(\theta)$ is almost linear on a log-log scale. The linear trend also represents a power law in the cost-quality tradeoff.
    \item \textbf{Phase 3} (red region) - The region of model parameters where the tradeoff has similar behavior to Phase 1. Models in this phase require significant investmest in $J(\theta)$ to obtain small improvements in $\mathcal{L}_{\mathcal{D}}$. 
\end{itemize}
We conjecture that models in Phase 1 encode a shared structure that most points in $\mathcal{D}$ conform to and is in some way \textit{easy} for the family of models to describe. 
Models in Phase 2 perform minute additions to this main structure to obtain better quality by encoding more data points.
By Phase 3, all data points in $\mathcal{D}$ that conform to the shared structure are captured by the model. 
The rest of the data will be nuisance variability that a model has to memorize, leading to the expensive tradeoff with respect to the cost $J(\theta)$.
The region near the boundary of Phase 2 and 3 is the region where the tradeoff between cost and quality becomes optimal.
This \textit{parameter phase} boundary becomes interesting in applications where resources are not constrained and the best possible model for a task is of interest.
In other words, this is the most likely region where the minimal sufficient statistic for a model family with respect to $\mathcal{D}$ can be found.
Figure \ref{fig:parameter_phases:comparison} shows how this three phase behavior is present in all the three traces we consider.

Now that we have established the general behavior of the parameter phases of the three cache miss rate traces, we will look at how we can compare the compression boundary curves of the traces.
Since the value of $\mathcal{L}_{\mathcal{D}}$ is the negative of the log likelihood of the models on the whole trace, this value for each trace cannot be compared directly.
We instead normalize $\mathcal{L}_{\mathcal{D}}$ by the length of the trace to obtain a comparable metric $\hat{\mathcal{L}}_{\mathcal{D}}$. 
Formally, the quantity we are interseted in comparing is defined as $\hat{\mathcal{L}}_{\mathcal{D}} = \frac{1}{N} \mathcal{L}_{\mathcal{D}}$ where $N$ is the number of points in the trace.
The comparison between $\hat{\mathcal{L}}_{\mathcal{D}}$ of the three traces is given in Figure \ref{fig:parameter_phases:comparison}. The figure shows that the cost at which the \textit{parameter phase} changes from 2 to 3 is in increasing order of the complexity of the \textit{program phase} behavior of each of the programs: swim-ref-swim < gcc-ref-cc15 < pmd-small-JikesRVM.
This interesting relation between complexity and resource requirements will be used to identify regions of program behavior that are complex for a model, which can be used to answer the second scientific question.

That second question is: What parts of a task require more complex models and how can we quantify and identify them?
To answer this question, we take models along the compression boundary that we found for each trace, in the increasing order of cost.
The likelihood achieved by each of the models at each point in the trace is taken to obtain a score of quality at that cost constraint.
Formally, we are interested in the value of $\mathcal{L}^t_{\mathcal{D}} = \log(f_{\theta}(x_t | x_{t-1}, x_{t-2}, ..., x_1))$ for each time index $t$ of the trace.
Since we have traces that have very large number of points, to visualize the likelihoods, we average these likelihoods in windows to obtain a local average likelihood.
This local average across time is plotted as a heat map with the x-axis as the time index, the y-axis as the cost of the model, and the colors indicating the magnitude of $\mathcal{L}^t_{\mathcal{D}}$.
The predictions of models at different costs are visualized alongside one another to obtain some idea about the type of predictions performed by models under different cost constraints.

We begin our discussion with swim-ref-swim as it has the least complex behavior among the three traces.
The lower plot of Figure \ref{fig:ll_heatmap:swim} shows the heat map of local likelihoods and Figure \ref{fig:cost_predictions:swim} shows the predictions of the corresponding models. 
It can be observed that at low cost regimes ($10$ to $2 \times 10^5$), models are able to learn only some parts of the trace and these parts occur periodically within the trace. 
The predictions that correspond to this regime seems to indicate constant predictions corresponding to the mode of the bins across the trace.
In the next regime ($2 \times 10^5$ to $2.5 \times 10^5$), part of the trace that is less complex to model also seems to be the low amplitude oscillations underlying the dynamic behavior of the trace.
As we consider models with even higher costs ($2.5 \times 10^5$ to $3 \times 10^5$), the corresponding predictions capture the periodic nature of the trace but fail to capture some of the variabilities.
These variabilities are captured as the cost bounds are increased and the predictions of high cost models show how it is possible to learn the general behavior of the whole trace except the rare peak that occurs.

We now consider gcc-ref-cc15 which unlike swim-ref-swim has multiple \textit{program phases} and has slightly complex behavior. The lower plot of Figure \ref{fig:ll_heatmap:gcc} shows the heat map of local average likelihoods and Figure \ref{fig:cost_predictions:gcc} shows the predictions of models under different resource constraints.
Unlike the swim-ref-swim trace, we observe that the improvement in quality of models obtained from relaxing cost constraints is different across the trace.
Although this is the case, quality improvements are not random, and interestingly they seems to track some of the \textit{program phases} that we observe within the trace.
Consider that gcc-ref-cc15 is divided into four arbitrary \textit{program phases} as in Figure \ref{fig:ll_heatmap:gcc}. It can be observed that the quality given by local likelihood average follows the trend: Phase 4 > Phase 2 > Phase 3 > Phase 1 for most cost levels. 
This shows how some \textit{program phases} may have more complex dynamical behavior than others. Thus it may be important to keep in mind, while modeling, the \textit{program phases} in which a resource constrained model will be effective.

Now we turn our attention to the more difficult pmd-small-JikesRVM trace, which does not exhibit long term periodic or phase-like behavior.
The local likelihood plot in Figure \ref{fig:ll_heatmap:pmd} shows how this complex behavior is reflected in the cost-quality tradeoff.
There are no distinct \textit{program phases} that can be observed in the trace or the local likelihood plots.
The predictions seem to improve only slowly as cost bounds are relaxed, but due to highly varying structure the cost required for a similar rate of improvement is much higher compared to the other two traces, even though the number of points in the trace is close to half that of the other two.

\section*{Conclusion and Future Work}

We introduced and elaborated a theoretical notion of resource constrained optimization problems and developed an empirical procedure to explore them.
We considered the tradeoff between quality and cost for specific resource constrained prediction models from the LSTM model family for the cache miss rate prediction problem.
We found that the tradeoff has a fairly regular trend for the traces we examined. We note how the point at which the process of learning a data set changes from learning the structure of task to memorization of the nuisance variability associated with it corresponds well with traditional notions of complexity of a data set.
In our evaluation, we showed how the procedure we developed can also be used to identify and quantify the complexity of dynamic behavior within a data set in addition to comparison among data sets.
Although we explored the scientific questions pertaining to resource constrained tasks, the practical applications of the above methods remain an open problem.
We anticipate that the procedure we developed, especially pertaining to identification and quantification of complexity of individual data points, can be applied to problems in curriculum learning, self-paced learning where complexity notions of individual data points within a data set are of interest.
%
Resource bounded Kolmogorov Complexity gives a new perspective on developing models for resource constrained problems based on $\beta$ sufficient statistics that opens up another avenue to create practical models for these problems.

\end{spacing}

\bibliography{citations} 
\bibliographystyle{apalike}
\end{document}